\documentclass[letterpaper, 10 pt, conference]{ieeeconf}  

\IEEEoverridecommandlockouts                              

\usepackage{times}

\usepackage{multicol}
\usepackage[bookmarks=true]{hyperref}

\usepackage{graphicx}
\usepackage{amsbsy}
\usepackage{amsmath}
\usepackage{amssymb}
\usepackage{multirow}
\usepackage{caption}

\usepackage[subrefformat=simple, labelformat=simple]{subcaption}
\usepackage{floatrow}

\usepackage{textcomp}
\usepackage{algorithm}
\usepackage{algpseudocode}
\usepackage{color}
\usepackage{xspace}
\usepackage{enumerate}

\title{
J-MOD\textsuperscript{2}: Joint Monocular Obstacle Detection and Depth Estimation
}

\author{Michele Mancini$^{1}$, Gabriele Costante$^{1}$, Paolo Valigi$^{1}$ and Thomas A. Ciarfuglia$^{1}$
\thanks{*We gratefully acknowledge the support of NVIDIA Corporation with the donation of the Titan X GPU used for this research.}
\thanks{$^{1}$All the authors are with the Department of Engineering, University of Perugia, via Duranti 93, Perugia Italy}
\thanks{{\tt\footnotesize \{thomas.ciarfuglia, paolo.valigi, gabriele.costante, michele.mancini\}@unipg.it}}
}

\def\LatexSettings{./settings}  
\def\include{./include}

\input{\LatexSettings/def_macros.tex} 

\begin{document}

\maketitle

\begin{abstract}
In this work, we propose an end-to-end deep architecture that jointly learns to detect obstacles and estimate their depth for MAV flight applications. Most of the existing approaches either rely on Visual SLAM systems or on depth estimation models to build 3D maps and detect obstacles. However, for the task of avoiding obstacles this level of complexity is not required. 
Recent works have proposed multi task architectures to both perform scene understanding and depth estimation. We follow their track and propose a specific architecture to jointly estimate depth and obstacles, without the need to compute a global map, but maintaining compatibility with a global SLAM system if needed.
The network architecture is devised to exploit the joint information of the obstacle detection task, that produces more reliable bounding boxes, with the depth estimation one, increasing the robustness of both to scenario changes. We call this architecture J-MOD\textsuperscript{2}.
We test the effectiveness of our approach with experiments on sequences with different appearance and focal lengths and compare it to SotA multi task methods that jointly perform semantic segmentation and depth estimation. In addition, we show the integration in a full system using a set of simulated navigation experiments where a MAV explores an unknown scenario and plans safe trajectories by using our detection model.
\end{abstract}

\section{Introduction}\label{sec:intro}

 \begin{figure}[t!]
 	\centering
 	\includegraphics[width=0.9\linewidth]{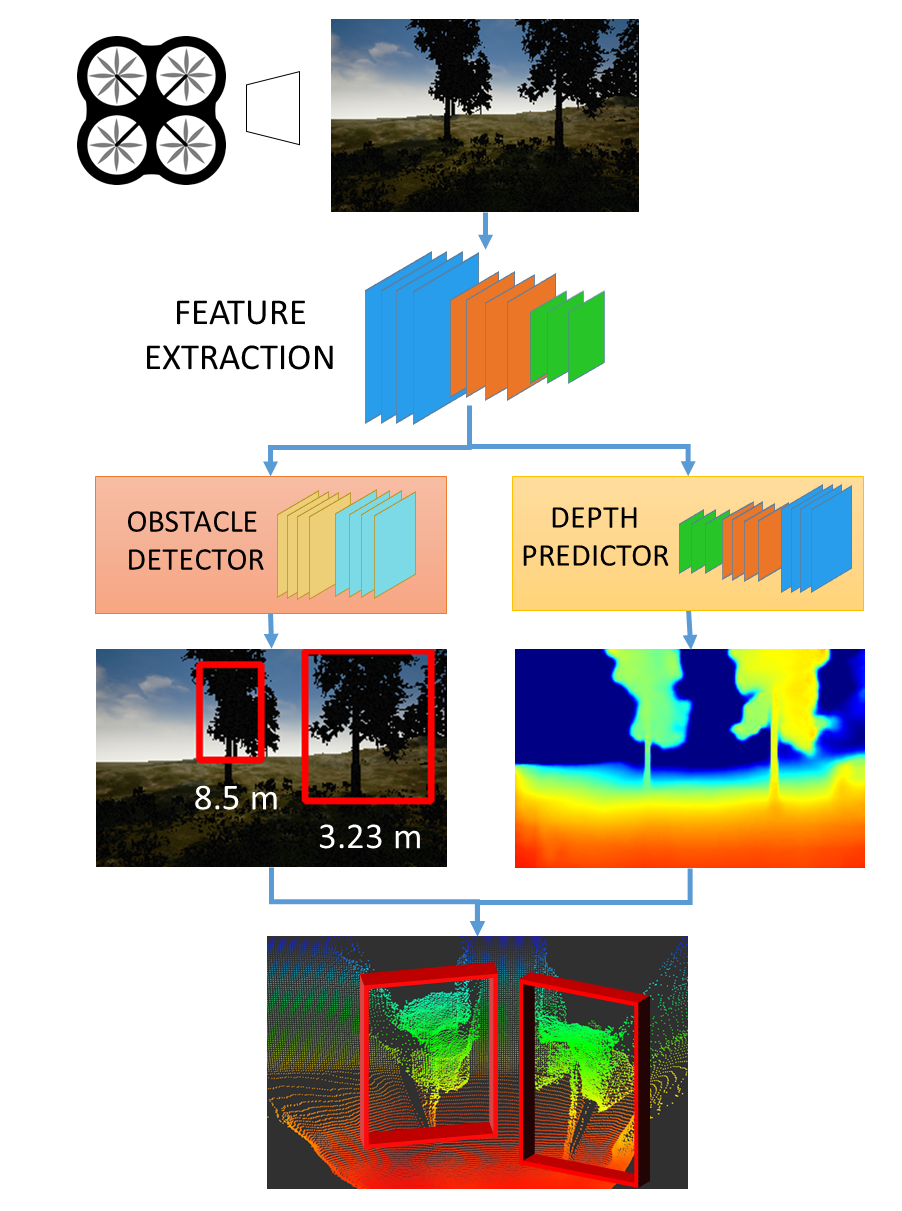}
 	\caption{\small Overview of the proposed system: the architecture is composed by two networks that perform different, but connected tasks: obstacle detection and pixel-wise depth estimation. The two task are jointly learned and the feature extraction layers are in common. Thus, the resulting model has increased accuracy in depth prediction because of the semantic information received from the detector. On the other hand, the detector learns a better representation of obstacles through depth estimation.\vspace{-1.6em}}
 	\label{fig:overview_depth}
 \end{figure}

Obstacle avoidance has been deeply studied in robotics due to its crucial role for vehicle navigation.
Recently, the demand for faster and more precise Micro Aerial Vehicle (MAV) platforms has put even more attention on it.
To safely execute aggressive maneuvers in unknown scenarios, the MAVs need a robust obstacle detection procedure.

Most fruitful approaches rely on range sensors, such as laser-scanner, stereo cameras 
or RGB-D cameras   \cite{grzonka2012fully, fraundorfer2012vision, bachrach2012estimation} to build 3D maps and compute obstacle-free trajectories.
However, their use results in an increased weight and power consumption, which is unfeasible for small MAVs. 
Furthermore, their sensing range is either limited by device characteristics (RGB-D and lasers) or by camera baselines (stereo cameras).

Monocular Visual SLAM (VSLAM) approaches address the above limitations by exploiting single camera pose estimation and 3D map 
reconstruction \cite{achtelik2014motion, scaramuzza2014vision, engel2014lsd, forster2014svo}.
Nevertheless, these advantages come with costs: the absolute scale is not observable (which easily results in wrong obstacle distance estimations); 
they fail to compute reliable 3D maps on low-textured environments; the 3D map updates are slow with respect to real-time requirements of fast manoeuvres. With careful tuning, these approaches can be used for obstacle avoidance.

At the same time there are other approaches that tackle the problem more specifically. In this respect, a step toward more robust obstacle detection has been made by monocular depth estimation methods based on Convolutional Neural Networks (CNNs) \cite{eigen2015predicting, mancini2017towards, yang2017obstacle}.
Compared to standard VSLAM strategies, these works train CNN-based model to quickly compute depth maps from single image, which allows for fast trajectory replanning.
However, as any data-driven approach, these depth models are biased with respect to appearance domains and camera intrinsics. Most of the CNN architectures so far proposed address the more general task of pixel-wise depth prediction and are not specifically devised for obstacle detection. However, recent works \cite{Cadena-RSS-16} \cite{jafari2017analyzing} have digressed from this trail, proposing multi task network architectures to jointly learning depth and some semantic property of the images. These works show that the mutual information is beneficial to both tasks.

Driven by the previous considerations, in this work we propose a novel CNN architecture that jointly learns the task of depth estimation and obstacle detection. The aim is to get the speed of detection of CNNs approaches, and, at the same time, make it more robust to scale and appearance changes using the joint learning of the depth distribution.
The combination of these two tasks gives them mutual advantages:
the depth prediction branch is informed with object structures, which result in more robust estimations. 
On the other hand, the obstacle detection model exploits the depth information to predict obstacle distance and bounding boxes more precisely.
Our approach is similar to \cite{Cadena-RSS-16} and \cite{jafari2017analyzing}, but is specifically devised for obstacle detection, and not generic scene understanding, in order to achieve more robustness to appearance changes. We show the comparison with these two aforementioned methods in the experimental part of the work.
We demonstrate the detection and depth estimation effectiveness of our approach in both publicly available and brand new sequences. 
In these experiments, we prove the robustness of the learned models in test scenarios that differ from the training ones with respect to focal length and appearance.
In addition, to demonstrate the detection advantages of the proposed detection system, we set up a full navigation avoidance system in a simulated environment with a MAV that detects obstacles and computes free trajectories as it explores the scene.

\section{Related Work} \label{sec:related}

The most straight-forward approaches to obstacle detection and depth estimation involve RGB-D or stereo cameras.
Unfortunately, these sensors suffer from limited range, in particular stereo systems, that require large baselines to achieve acceptable performances \cite{nous2016performance}.
For example, some authors explored push-broom stereo systems on fixed-wing, high speed MAVs \cite{barry2015pushbroom}. However, these approaches require too large baselines for small rotary wing MAVs. In addition, while short-range estimations still allows safe collision avoidance, it sets an upper bound to the robot's maximum operative speed.
For all these reasons the study of alternative systems based on monocular cameras becomes relevant. Even with the limitation of monocular vision, our method can detect and localize obstacles up to 20 meters and compute dense depth maps up to 40 meters with a minor payload and space consumption. 

Monocular obstacle detection can be achieved by dense 3D map reconstruction via SLAM or Structure from Motion (SFM) based procedures \cite{engel2014lsd}, \cite{pizzoli2014remode}, \cite{alvarez2016collision}. 
These systems perform a much more complex task though, and usually fail at high speeds, since they reconstruct the environment from frame to frame triangulation.
In addition, with standard geometric monocular systems it is not possible to recover the absolute scale of the objects, without using additional information. In \cite{Geiger2011IV} the scale is recovered using the knowledge of the camera height from the ground plane, while  \cite{frost2016object} uses a inference based method on the average size of objects that frequently appear in the images (e.g. cars), then optimize to the whole trajectory. The lack of knowledge of the scale makes the obstacle avoidance a difficult task. For this reason, some approaches exploit optical information to detect proximity of obstacles from camera, or, similarly, detect traversable 
space, or use hand-crafted image features \cite{ross2013learning}, \cite{DaftryZKDMBH16robust}, \cite{beyeler2009vision}, \cite{bills2011autonomous}, \cite{mori2013first}.



However, recently proposed deep learning-based solutions have shown robustness to the aforementioned issues. These models produce a dense 3D representation of the environment from a single image, exploiting the knowledge acquired through training on large labeled datasets, both real-world and synthetic \cite{eigen2014depth}, \cite{eigen2015predicting}, \cite{liu2015learning}, \cite{mancini2017towards}. 
A few of these methods have been recently tested in obstacle detection and autonomous flight applications. In \cite{chakravarty2017cnn}, the authors fine-tune on a self-collected dataset the depth estimation model proposed by \cite{eigen2014depth} and use it for path planning. In \cite{yang2017obstacle} the authors exploit depth and normals estimations of a deep model presented in \cite{eigen2015predicting} as an intermediate step to train an visual reactive obstacle avoidance system. More recently, \cite{yang2017obstacle} proposed a similar approach, regressing avoidance paths directly from monocular 3D depth maps.

However, the aforementioned methods solve the task of depth estimation and from it derive the obstacle map. Another set of approaches use semantic knowledge to strengthen the detection task. On this line the works of \cite{KendallGC17multi}, \cite{Cadena-RSS-16} and \cite{jafari2017analyzing} train a multi task architecture for semantic scene understanding that is reinforced by the joint learning of a depth estimation task. However, these methods show better performances on classes such as "ground" or "sky". Our intuition is that current depth estimators overfit their predictions on these classes, as they tend to have more regular texture and geometric structures.
On the contrary, in robotic applications we want to train detection models to be as accurate as possible when estimating obstacle distances. 

Following this multi task approaches, we propose a novel solution to the problem by jointly training a model for depth estimation and obstacle detection. While each task's output comes from independent branches of the network, feature extraction from their common RGB input is shared for both targets. This choice improves both depth and detection estimations compared to single task models, as shown in the experiments. An approach similar to ours, applied to 3D bounding box detection, is presented in \cite{mousavian20173d}, where the authors train a three-loss model, sharing the feature extraction layers between the tasks.

In our system the obstacles bounding box regression part is obtained modifying the architecture of \cite{redmon2016you} making it fully convolutional. This allows for multiple bounding box predictions with a single forward pass. In addition, we also ask the obstacle detector to regress the average depth and the corresponding estimate variance of the detected obstacles.

Depth estimation is devised following the architecture of \cite{mancini2017towards}, improved by taking into account the obstacle detection branch.
In particular, we correct the depth predictions by using the mean depth estimates computed by the obstacle detection branch to achieve robustness with respect to appearance changes.
We prove the benefits of this strategy by validating the model in test sequences with different focal length and scene appearance. We compare our method to the ones of \cite{Cadena-RSS-16} and \cite{jafari2017analyzing}, showing a considerable increase of performances over these two baselines.

\begin{figure*}[ht!]
	\centering
	\includegraphics[scale=0.3]{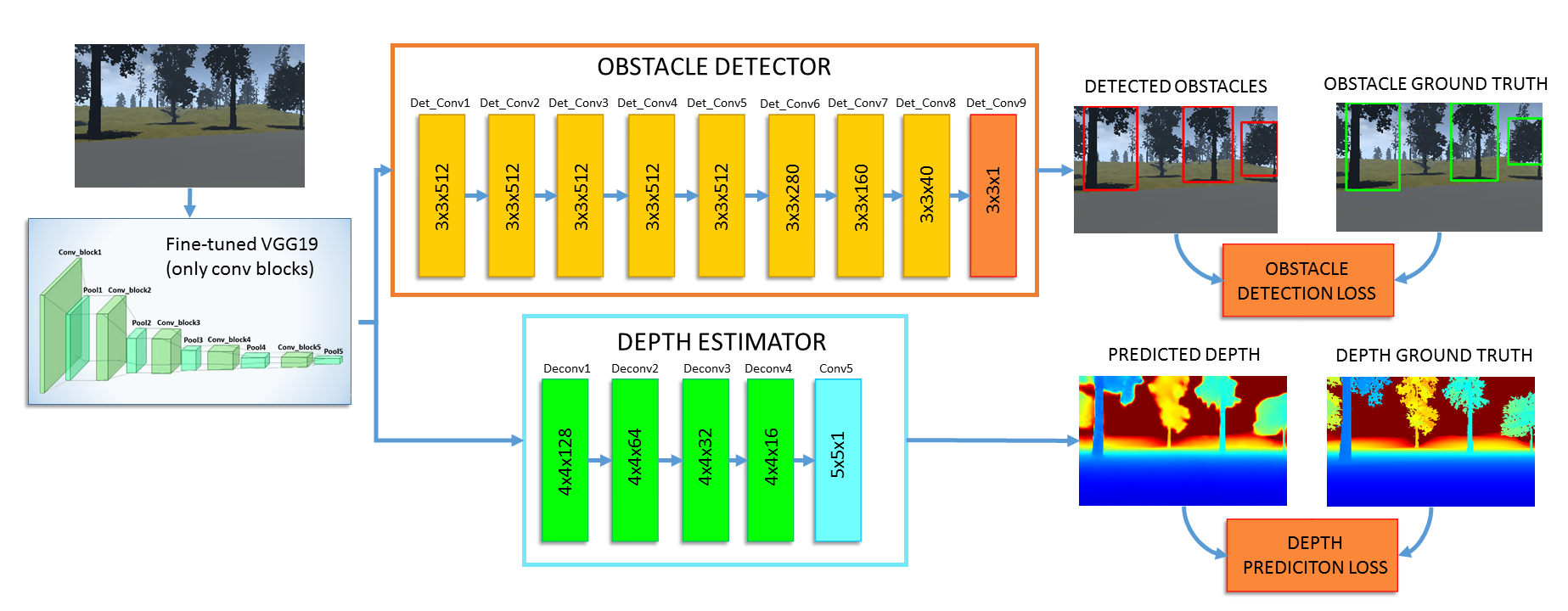} 
	\vspace{-1.2em}
	\caption{\small Architecture of J-MOD$^{2}$. Given an RGB input, features are extracted by the VGG19 module and then fed into the depth estimation and obstacle detection branches to produce dense depth maps and osbtacles bounding boxes.\label{fig:architecture} \vspace{-1.6em}}
	
\end{figure*}
\section{Network Overview}
Our proposed network is depicted in Figure \ref{fig:architecture}. Given an $256 \times 160$ RGB input, features are extracted with a fine-tuned version of the VGG19 network pruned of its fully connected layers \cite{simonyan2014very}. VGG19 weights are initialized on the image classification task on the ImageNet dataset. Features are then fed to two, task-dependent branches: a depth prediction branch and a obstacle detector branch. The former is composed by 4 upconvolution layers and a final convolution layer which outputs the predicted depth at original input resolution. This branch, plus the VGG19 feature extractor, is equivalent to the fully convolutional network proposed in \cite{mancini2017towards}. 
We optimize depth prediction on the following loss:
\begin{equation}
\begin{split}
	L_{depth} = & \frac{1}{n}\sum_{i} d_{i}^{2} - \frac{1}{2n^{2}}(\sum_{i} d_{i})^{2}\\  
	            & + \frac{1}{n}\sum_{i} [\nabla_{x}{ D_{i}} + \nabla_{y}{D_{i}}] \cdot N_{i}^{*}
\end{split}
\end{equation} 
where $d_{i}=\log D_{i}-\log D_{i}^{*}$, $D_{i}$ and $D_{i}^{*}$ are respectively the predicted and ground truth depths at pixel $i$, $N_{i}^{*}$ is the ground truth 3D surface normal, and $\nabla_{x}{ D_{i}}$, $\nabla_{y}{D_{i}}$ are the horizontal and vertical predicted depth gradients. 
While the first two terms correspond to the scale invariant log RMSE loss introduced in \cite{eigen2014depth}, the third term enforces orthogonality between predicted gradients and ground truth normals, aiming at preserving geometrical coherence. With respect to the loss proposed in \cite{eigen2015predicting}, that introduced a L2 penalty on gradients to the scale invariant loss, our loss performs comparably in preliminary tests.

The obstacle detection branch is composed by 9 convolutional layer with Glorot initialization. The detection methodology is similar to the one presented in \cite{redmon2016you}: the input image is divided into a $8 \times 5$ grid of square-shaped cells of size $32 \times 32$ pixels. For each cell, we train a detector to estimate:
\begin{itemize}
\item The $(x,y)$ coordinates of the bounding box center
\item The bounding box width $w$ and height $h$
\item A confidence score $C$
\item The average distance of the detected obstacle from the camera $m$ and the variance of its depth distribution $v$
\end{itemize}
The resulting output has a $40 \times 7$ shape. At test time, we consider only predictions with a confidence score over a certain threshold. 
We train the detector on the following loss:

\begin{equation}
\begin{split}
L_{det} = & \quad \lambda_{coord}\sum_{i=0}^{N}[(x_{i} - x_{i}^*)^{2} + (y_{i} - y_{i}^{*})^2] \\
& + \lambda_{coord}\sum_{i=0}^{N}[(w_{i} - w_{i}^*)^{2} + (h_{i} - h_{i}^{*})^2] \\
& + \lambda_{obj}\sum_{i=0}^{N}(C_{i} - C_{i}^*)^{2} + \lambda_{noobj}\sum_{i=0}^{N}(C_{i} - C_{i}^*)^{2}\\
& + \lambda_{mean}\sum_{i=0}^{N}(m_{i} - m_{i}^*)^{2} + \lambda_{var}\sum_{i=0}^{N}(v_{i} - v_{i}^*)^{2}\\
\end{split}
\end{equation}

where we set $\lambda_{coord} = 0.25$, $\lambda_{obj}= 5.0$, $\lambda_{noobj} = 0.05$, $\lambda_{mean} = 1.5$, $\lambda_{var} = 1.25$.
Our network is trained simultaneously on both tasks. Gradients computed by each loss are backpropagated through their respective branches and the shared VGG19 multi-task feature extractor.
\subsection{Exploiting detection to correct global scale estimations} \label{sec:correction}
The absolute scale of a depth estimation is not observable from a single image. However, learning-based depth estimators are able to give an accurate guess of the scale under certain conditions.
While training, these models implicitly learn domain-specific object proportions and appearances. This helps the estimation process in giving depth maps with correct absolute scale.
As the relations between object proportions and global scale in the image strongly depend on camera focal length, at test time the absolute scale estimation are strongly biased towards the training set domain and its intrinsics.
For these reasons, when object proportions and/or camera parameters change from training to test, scale estimates quickly degrade. Nonetheless, if object proportions stay roughly the same and only camera intrinsics are altered at test time, it is possible to employ some recovery strategy. If the size of a given object is known, we can analytically compute its distance from the camera and recover the global scale for the whole depth map. For this reason, we suppose that the obstacle detection branch can help recovering the global scale when intrinsics change. We hypothesize that, while learning to regress obstacles bounding boxes, a detector model implicitly learns sizes and proportions of objects belonging to the training domain. We can then evaluate estimated obstacle distances from the detection branch and use them as a tool to correct dense depth estimations. Let $m_{j}$ be the average distance of the obstacle $j$ computed by the detector, $\hat{D}_{j}$ the average depth estimation within the $j$-th obstacle bounding box, $n_o$ the number of estimated obstacles, then we compute the correction factor $k$ as:
\begin{equation}
k = \frac{\frac{1}{n_o}\sum_{j}^{n_o}m_{j}}{\frac{1}{n_o}\sum_{j}^{n_o}\hat{D}_{j}}
\end{equation}
Finally, we calculate the corrected depth at each pixel $i$ as $\tilde{D}_i = kD_i$. To validate our hypothesis, in Section \ref{sec:unreal_test} we test on target domains with  camera focal lengths that differ from the one used for training.

\section{Experiments}\label{sec:exp}

\subsection{Datasets}
\subsubsection{UnrealDataset}\label{subsec:unreal}
\begin{figure}[h]
	\centering
	\includegraphics[scale=0.32]{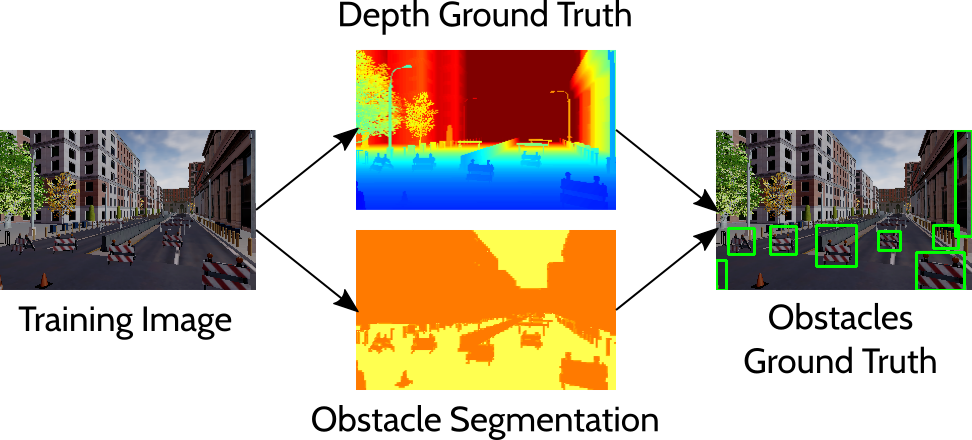} 
	\vspace{0.5em}
	\caption{\small Given depth and segmentation ground truth, we compute obstacle bounding boxes for each training image. We evaluate only obstacles in a 20 meters range. \label{fig:obstacle_segmentation_gt_example}}
	\vspace{-1.0em}
\end{figure}

\begin{figure*}[h]
\centering
  \ffigbox{}
  {
    \CommonHeightRow
    {
      \begin{subfloatrow}[5]
      \hspace{-0.8em}
    \ffigbox[\FBwidth]
    {\includegraphics[height=\CommonHeight]{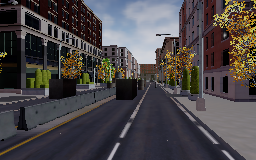}}
    {\vspace{0.5 em}}
    \hspace{-1.5em}
    \ffigbox[\FBwidth]
    {\includegraphics[height=\CommonHeight]{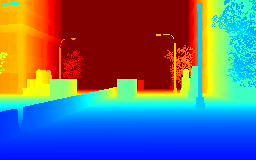}}
    {\vspace{0.5em}}
    \hspace{-1.5em}
    \ffigbox[\FBwidth]
    {\includegraphics[height=\CommonHeight]{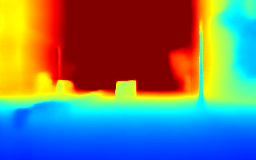}}
    {\vspace{0.5em}}
    \hspace{-1.5em}
    \ffigbox[\FBwidth]
    {\includegraphics[height=\CommonHeight]{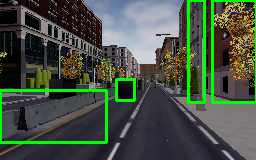}}
    {\vspace{0.5em}}
    \hspace{-1.5em}
    \ffigbox[\FBwidth]
    {\includegraphics[height=\CommonHeight]{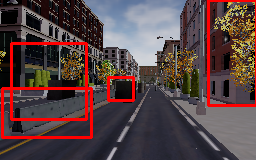}}
    {\vspace{0.5em}}
      \end{subfloatrow}
    } 
    
    \CommonHeightRow
    {
      \begin{subfloatrow}[5]
      \hspace{-0.8em}
    \ffigbox[\FBwidth]
    {\includegraphics[height=\CommonHeight]{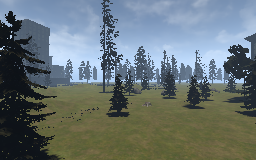}\caption{\footnotesize RGB Input}
    \label{fig:unreal_example}}
    {\vspace{-0.5 em}}
    \hspace{-1.5em}
    \ffigbox[\FBwidth]
    {\includegraphics[height=\CommonHeight]{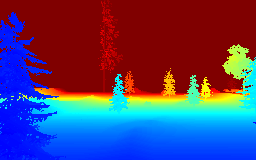}\caption{\footnotesize Depth GT}}
    {\vspace{-0.5em}}
    \hspace{-1.5em}
    \ffigbox[\FBwidth]
    {\includegraphics[height=\CommonHeight]{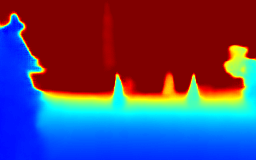}\caption{\footnotesize Depth Estimation}}
    {\vspace{-0.5em}}
    \hspace{-1.5em}
    \ffigbox[\FBwidth]
    {\includegraphics[height=\CommonHeight]{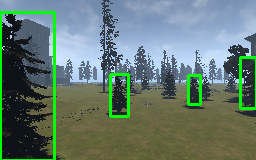}\caption{\footnotesize Obstacle GT}}
    {\vspace{-0.5em}}
    \hspace{-1.5em}
    \ffigbox[\FBwidth]
    {\includegraphics[height=\CommonHeight]{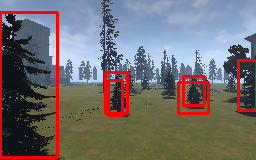}\caption{\footnotesize Detected Obstacles}}
    {\vspace{-0.5em}}
      \end{subfloatrow}
    }

    \vspace{-1.0em}
      \caption{\small J-MOD$^{2}$ qualitative results on the UnrealDataset. \vspace{-1.6em}}  
       \label{fig:unreal_results}
    }
\end{figure*}

%
%
UnrealDataset is a self-collected synthetic dataset that comprises of more than 100k images and 21 sequences collected in a bunch of highly photorealistic urban and forest scenarios with Unreal Engine and the AirSim plugin \cite{shah2017airsim}, which allows us to navigate a simulated MAV inside any Unreal scenarios. The plugin also allows us to collect MAV's frontal camera RGB images, ground truth depth up to 40 meters and segmentation labels. Some samples are shown in Figure \ref{fig:unreal_example}. We postprocess segmentation labels to form a binary image depicting only two semantic classes: obstacle and non-obstacle by filtering these data with corresponding depth maps, we are finally able to segment obstacles at up to 20 meters from the camera and get ground truth labels for the detecFtion network branch (Fig. \ref {fig:obstacle_segmentation_gt_example}). MAV's frontal camera has a horizontal field of view of 81,5 degrees.
\subsubsection{Zurich Forest Dataset}\label{subsec:zurich_data}
Zurich Forest Dataset consist of 9846 real-world grayscale images collected with a hand-held stereo camera rig in a forest area. Ground truth depth maps are obtained for the whole dataset through semi-global stereo matching \cite{hirschmuller2008stereo}. We manually draw 357 bounding boxes on a subset of 64 images to provide obstacle ground truth and evaluate detection in a real-world scenario.

\begin{table*}[h]
\vspace{0.8em}
  \centering
  \caption{\small Results on the UnrealDataset. For the depth estimation task we report full depth map RMSE and scale invariant errors, obstacle-wise depth and detection branches statistics (mean/variance) estimation errors and detector's IOU, precision and recall.\vspace{-1.6em}}
  
  \label{tab:unreal_results}
  \begin{scriptsize}

  \begin{tabular}{|c||c|c|c|c|c|c||c|}
    \hline
     &  DEPTH \cite{mancini2017towards} &  DETECTOR &  EIGEN \cite{eigen2015predicting} &
      FULL-MAE \cite{Cadena-RSS-16} & JRN \cite{jafari2017analyzing} &  J-MOD$^{2}$ & \\
	\hline
	 RMSE Full Depth Map 				&   3.653 & - &  3.785 &  7.566 &   7.242 &  \textbf{3.473} &  Lower\\
	 Sc.Inv RMSE Full Depth Map  		&   0.042 & - &  0.043& 0.124 &  0.110 &  \textbf{0.036} &   is \\
	 Depth RMSE on Obs.(Mean/Var) 	&  1.317 / 37.124 & - & 1.854 / 50.71 & 5.355/180.67 &  2.938 / 87.595 & \textbf{ 1.034 / 29.583} & better \\
	 Detection RMSE on Obs.(Mean/Var) & - & 2.307/ 59.407 & - & - & - &  \textbf{ 1.754 / 46.006} & \\	\hline
	\hline
	 Detection IOU & - &  63.11\% & - &  32.58\% &  44.19\% &  \textbf{ 66.58\%} &  Higher\\
	 Detection Precision & - &  72.15\% & - &  75.53\%  &  54.37\% &   \textbf{78.64\%} &  is \\
	 Detection Recall & - &  90.05\% & - &  44.38\% & 49.55\% &   \textbf{90.85\%} &  better \\
	\hline
  \end{tabular}
    \end{scriptsize}
\end{table*}

\subsection{Training and testing details}
As baselines, we compare J-MOD$^{2}$ with:
\begin{itemize}
\item The depth estimation method proposed in \cite{mancini2017towards}.
\item Our implementation of the multi-scale Eigen's model \cite{eigen2015predicting}.
\item A simple obstacle detector, consisting of our proposed model, trained without the depth estimation branch.
\item Our implementation of the multi-modal autoencoder (later referred as Full-MAE) proposed by Cadena et al. \cite{Cadena-RSS-16}.
\item Our implementation of the joint refinement network (later referred as JRN) proposed by Jafari et al. \cite{jafari2017analyzing}.
\end{itemize} 
We train J-MOD$^{2}$ and all the baseline models on 19 sequences of the UnrealDataset. We left out sequences 09 and 14 for testing. All the approaches have been trained on a single NVIDIA Titan X GPU. Training is performed with Adam optimizer by setting a learning rate of 0.0001 until convergence. The segmentation tasks for the Full-MAE and the JRN baselines are trained to classify two classes: "obstacle" and "not obstacle". The JRN is trained to fuse and refine depth estimations from our implementation of \cite{eigen2015predicting} with segmentation estimates from the SotA segmentation algorithm of Long et al. \cite{long2015fully}, as suggested by the authors, with the latter retrained on the 2-class segmentation problem of the UnrealDataset.

At test time, all baseline methods are tested using only RGB inputs.
For both methods, we then infer obstacle bounding boxes from their depth and segmentation estimates applying the same procedure described in Figure \ref{fig:obstacle_segmentation_gt_example}, allowing direct comparison with our method.
All the approaches are tested on the test sequences of the UnrealDataset and on the whole Zurich Forest Dataset. Note that, while testing on the latter, we do not perform any finetuning for both our method and the baselines. 

At runtime, estimations require about 0.01 seconds per frame on a NVIDIA Titan X GPU. We also test J-MOD$^{2}$ on a NVIDIA TX1 board, to evaluate its portability on a on-board embedded system, measuring an average forward time of about 0.28 seconds per frame.
The code for J-MOD$^{2}$ and all the baseline methods is available online{\footnote{\url{http://isar.unipg.it/index.php?option=com_content&view=article&id=47&catid=2&Itemid=188}}}

To evaluate the depth estimator branch performance, we compute the following metrics:
\begin{itemize}
\item Linear RMSE and Scale Invariant Log RMSE ($ \frac{1}{n}\sum_{i} d_{i}^{2} - \frac{1}{n^{2}}(\sum_{i} d_{i})^{2}$, with $d_{i} = \log y_{i}-\log y_{i}^{*}$) on the full depth map.
\item Depth RMSE on Obstacles (Mean/Variance): For each ground truth obstacle, we compute its depth statistics (mean and variance) and we compare them against the estimated ones by using linear RMSE.
\end{itemize}
For the detector branch, we compute the following metrics: 
\begin{itemize}
\item Detection RMSE on Obstacles (Mean/Variance):For each detected obstacle, we compare its estimated obstacle depth statistics (mean and variance) with the closest obstacle ones by using linear RMSE.
\item Intersection Over Union (IOU)
\item Precision and Recall.
\end{itemize}

\subsection{Test on UnrealDataset} \label{sec:unreal_test}
We report results on the UnrealDataset on Table \ref{tab:unreal_results}. For \cite{mancini2017towards} and \cite{eigen2015predicting} we report results only on depth-related metrics, as they do not perform any detection. Results confirm how J-MOD$^{2}$ outperforms all the other baselines in all metrics, corroborating our starting claim: object structures learned by the detector branch improve obstacles depth estimations of the depth branch. At the same time, localization and accuracy of the detected bounding boxes improve significantly compared to our single-task obstacle detector. Qualitative results are shown on Figure \ref{fig:unreal_results}. 

\begin{table*}[ht]
\vspace{0.8em}
  \centering
  \caption{\small Results of J-MOD$^{2}$ on the sequence-20 of the UnrealDataset on different-sized central crops. For each crop, we report in bold the better estimation between unchanged (labeled as NoCor) and corrected depths (labeled as WithCor).}
 
  \label{tab:crop_results}
  \begin{scriptsize}
  \begin{tabular}{|c||c|c|c|c|c|c|c|c|c|c|}
    \hline

    \multirow{2}{*} & \multicolumn{2}{|c|}{ ORIGINAL SIZE} &  \multicolumn{2}{|c|}{ CROP 230X144} &  \multicolumn{2}{|c|}{ CROP 204X128}&  \multicolumn{2}{|c|}{ CROP 154X96}&  \multicolumn{2}{|c|}{ CROP 128X80}\\
    \cline{2-11}
     {} &  Cor &  NoCor &  Cor &  NoCor &  Cor &  NoCor &  Cor &  NoCor &  Cor &  NoCor\\
    \hline
     RMSE Full Depth Map &  \textbf{2.179} &  2.595 &  \textbf{2.632} &  3.042 &  4.052 &  \textbf{3.991} &  8.098 &  \textbf{6.234} &  10.825 &  \textbf{8.045}  \\
     Sc. Inv RMSE Full Depth Map &  \textbf{0.096} &  0.115 &  \textbf{0.121} &  0.134 &  0.173 &  \textbf{0.164} &  0.274 &  \textbf{0.217} &  0.305 &  \textbf{0.250} \\
     Depth RMSE on Obs.(Mean) &  \textbf{0.185} &  0.676 &  \textbf{1.293} &  1.458 &  2.465 &  \textbf{2.219} &  4.865 &  \textbf{3.583} &  6.148 &  \textbf{4.485} \\
   \hline   
    Detector RMSE on Obs.(Mean) & \multicolumn{2}{|c|}{ 0.404} &  \multicolumn{2}{|c|}{ 1.079} &  \multicolumn{2}{|c|}{ 1.998}&  \multicolumn{2}{|c|}{ 4.124}&  \multicolumn{2}{|c|}{ 5.450}\\
    \hline
  \end{tabular}
  \vspace{-0.8em}
  \end{scriptsize}
\end{table*}

To validate our proposed depth correction strategy introduced in Section \ref{sec:correction}, we simulate focal length alterations by cropping and upsampling a central region of the input images of the UnrealDataset. We evaluate performances on different sized crops of images on the sequence-20, comprising of more than 7700 images. It is worth mentioning that sequence-20 is one of the training sequences. We choose to stage this experiment on a training sequence rather than on a test one to minimize appearance-induced error and make evident the focal-length-induced error. We report results on Table \ref{tab:crop_results}. When no crop is applied, camera intrinsics are unaltered and appearance-induced error is very low, as expected. As correction is applied linearly on the whole depth map, when scale-dependant error is absent or low, such correction worsen estimations by $19\%$ on non-cropped images. A $230 \times 144$ crop simulates a slightly longer focal length. All metrics worsen, as expected, and correction still cause a $15\%$ higher RMSE error. When $204 \times 128$ crops are evaluated, correction starts to be effective, improving performances by $1,45\%$ with respect to the non-corrected estimation. On $154 \times 96$ crops, correction leads to a $23\%$ improvement. On $128 \times 80$ crops, correction improves performance by $25\%$. We also observe how the detection branch outperforms the depth estimation branch on obstacle distance evaluation as we apply wider crops to the input. This results uphold our hypothesis that detection branch is more robust to large mismatches between training and test camera focal lengths and can be used to partially compensate the induced absolute scale estimation deterioration.

\subsection{Test: Zurich Forest Dataset}\label{sec:zurich}
\begin{figure*}[ht!]
\centering
  \ffigbox{}
  {
    \CommonHeightRow
    {
      \begin{subfloatrow}[5]
      \hspace{-0.8em}
    \ffigbox[\FBwidth]
    {\includegraphics[height=\CommonHeight]{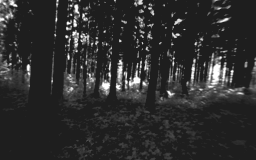}}
    {\vspace{0.5 em}}
    \hspace{-1.5em}
    \ffigbox[\FBwidth]
    {\includegraphics[height=\CommonHeight]{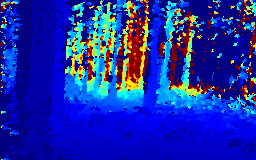}}
    {\vspace{0.5em}}
    \hspace{-1.5em}
    \ffigbox[\FBwidth]
    {\includegraphics[height=\CommonHeight]{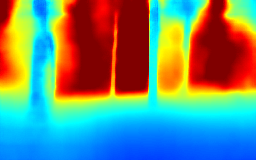}}
    {\vspace{0.5em}}
    \hspace{-1.5em}
    \ffigbox[\FBwidth]
    {\includegraphics[height=\CommonHeight]{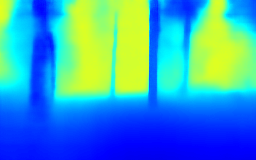}}
    {\vspace{0.5em}}
    \hspace{-1.5em}
    \ffigbox[\FBwidth]
    {\includegraphics[height=\CommonHeight]{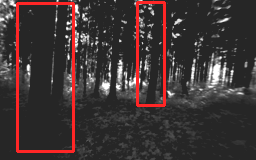}}
    {\vspace{0.5em}}
      \end{subfloatrow}
    } 
    
    \CommonHeightRow
    {
      \begin{subfloatrow}[5]
      \hspace{-0.8em}
    \ffigbox[\FBwidth]
    {\includegraphics[height=\CommonHeight]{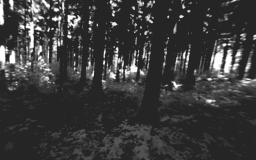}\caption{\footnotesize RGB Input}}
    {\vspace{-0.5 em}}
    \hspace{-1.5em}
    \ffigbox[\FBwidth]
    {\includegraphics[height=\CommonHeight]{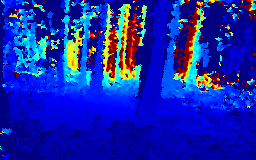}\caption{\footnotesize Depth GT}}
    {\vspace{-0.5em}}
    \hspace{-1.5em}
    \ffigbox[\FBwidth]
    {\includegraphics[height=\CommonHeight]{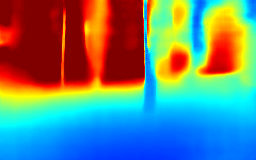}\caption{\footnotesize Non Corrected Depth Estimation}}
    {\vspace{-0.5em}}
    \hspace{-1.5em}
    \ffigbox[\FBwidth]
    {\includegraphics[height=\CommonHeight]{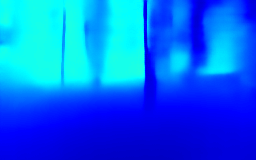}\caption{\footnotesize Corrected Depth Estimation}}
    {\vspace{-0.5em}}
    \hspace{-1.5em}
    \ffigbox[\FBwidth]
    {\includegraphics[height=\CommonHeight]{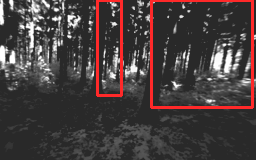}\caption{\footnotesize Detected Obstacles}}
    {\vspace{-0.5em}}
      \end{subfloatrow}
    } 
    \vspace{-0.3em}
      \caption{ \small J-MOD$^{2}$ qualitative results on the Zurich Forest Dataset.\vspace{-0.4em}}  
       \label{fig:zurich_results}
    }
\end{figure*}

\begin{table*}[ht]
\vspace{0.8em}
  \centering
  
  \caption{\small Results on the Zurich Forest Dataset. Metrics marked with a $^{*}$ symbol are evaluated on a subset of 64 images with ground truth bounding boxes. \vspace{-1.0em}}
 \begin{scriptsize}
  \label{tab:zurich}
  \begin{tabular}{|c||c|c|c|c|c|c|c|c|c|c|c|c|}
    \hline

    \multirow{2}{*} & \multicolumn{2}{|c|}{ DEPTH \cite{mancini2017towards}} & \multicolumn{2}{|c|}{ DETECTOR} & \multicolumn{2}{|c|}{ EIGEN \cite{eigen2015predicting}} &  \multicolumn{2}{|c|}{ FULL-MAE \cite{Cadena-RSS-16} } &  \multicolumn{2}{|c|}{ JRN \cite{jafari2017analyzing}} & \multicolumn{2}{|c|}{ J-MOD$^{2}$}\\
    \cline{2-13}
    \footnotesize {} &  Cor &  NoCor &  Cor &  NoCor &  Cor &  NoCor &  Cor &  NoCor &  Cor &  NoCor &  Cor &  NoCor\\
    \hline
      RMSE & - & 12.421 & - & - & - & 14.640& - & 17.581 & - & 10.114 & \textbf{9.009} &   12.569 \\
      Sc. Inv RMSE & - & 0.873 & - & - & - & 1.025 & - & 1.711 & - & 0.702 & \textbf{0.429}& 0.954 \\
      Depth RMSE on Obs.(Mean)$^{*}$ & - & \textbf{4.378} & - & - & - & 8.060 & - & 10.488 & - & 4.783 & 4.510 & 4.847 \\
    \hline
      Detector RMSE on Obs.(Mean)$^{*}$ & \multicolumn{2}{|c|}{ -} &  \multicolumn{2}{|c|}{ 6.277 } &  \multicolumn{2}{|c|}{  -}&  \multicolumn{2}{|c|}{  -} & \multicolumn{2}{|c|}{  -} & \multicolumn{2}{|c|}{\textbf{3.702}}\\
    
       Detector IOU$^{*}$ & \multicolumn{2}{|c|}{ -} &  \multicolumn{2}{|c|}{  14.4\% } &  \multicolumn{2}{|c|}{  -}&  \multicolumn{2}{|c|}{  2.13\%} & \multicolumn{2}{|c|}{  9.19\%} & \multicolumn{2}{|c|} {\textbf{26.32\%}}\\
     
       Detector Precision$^{*}$ & \multicolumn{2}{|c|}{  -} &  \multicolumn{2}{|c|}{  25.32\% } &  \multicolumn{2}{|c|}{ -}&  \multicolumn{2}{|c|}{  11.4\%} & \multicolumn{2}{|c|}{  13.18\%} & \multicolumn{2}{|c|}{ \textbf{48.36\%}}\\
      
       Detector Recall$^{*}$ & \multicolumn{2}{|c|}{  -} &  \multicolumn{2}{|c|}{  10.80\% } &  \multicolumn{2}{|c|}{  -}&  \multicolumn{2}{|c|}{  1.12\%} & \multicolumn{2}{|c|}{  6.72\%} & \multicolumn{2}{|c|}{  \textbf{20.49\%}}\\
    \hline
  \end{tabular}
  \vspace{-0.8em}
 \end{scriptsize}
\end{table*}

Intrinsic parameters of this dataset do not match the UnrealDataset ones, causing large scale-induced errors. Therefore, we can evaluate the performance of J-MOD$^{2}$ corrected depth, as introduced in Section \ref{sec:correction}. In addition, we also evaluate the performances of all the other baselines. Depth metrics (Linear RMSE and Scale Invariant MSE) refer to the whole dataset, while all the other metrics refer to the labelled subset, as described in Section \ref{subsec:zurich_data}. Results are reported on Table \ref{tab:zurich}. J-MOD$^{2}$ outperforms all baselines in almost all metrics, which suggests improved generalization capabilities. Furthermore, we show how the correction factor improves J-MOD$^{2}$ depth estimation by about 28\% on the RMSE metric. 

\subsection{Qualitative analysis of the multi-task interaction}

  \begin{figure}[ht!]
\centering
 \begin{subfigure}[text]{0.49\columnwidth}
    {\includegraphics[width=0.99\columnwidth]{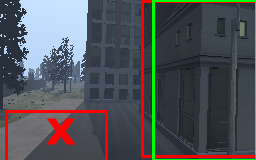}}
     {\caption{\footnotesize Single-task detector}\label{fig:det_analysis_a}}
  \end{subfigure}
  \begin{subfigure}[text]{0.49\columnwidth}
    {\includegraphics[width=0.99\columnwidth]{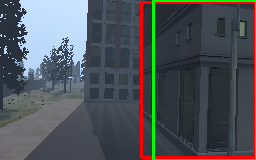}}
     {\caption{\footnotesize J-MOD$^{2}$}\label{fig:det_analysis_b}}
  \end{subfigure}
  
    \begin{subfigure}[text]{0.49\columnwidth}
    {\includegraphics[width=0.99\columnwidth]{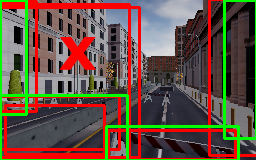}}
    {\caption{\footnotesize Single-task detector}\label{fig:det_analysis_c}}
  \end{subfigure}
  \begin{subfigure}[text]{0.49\columnwidth}
    {\includegraphics[width=0.99\columnwidth]{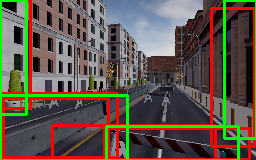}}
    {\caption{\footnotesize J-MOD$^{2}$}\label{fig:det_analysis_d}}
  \end{subfigure}

  \caption{\small For each row, we compare J-MOD$^{2}$ obstacle detections with the detector-only architecture. Ground truth bounding boxes are reported in green, predictions in red. In the first example (first row), the single-task detector erroneously detects a false obstacle on the ground. Similarly, in the second example (second row), the single-task wrongly considers the whole building on the left as an obstacle while only its closest part is an immediate threat for robot navigation.\vspace{-1.6em}}  
       \label{fig:det_analysis}
\end{figure}

  \begin{figure}[ht!]
\centering
 \begin{subfigure}[text]{0.24\columnwidth}
    {\includegraphics[width=0.99\columnwidth]{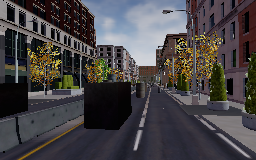}}
     {\caption{\footnotesize RGB image}\label{fig:depth_analysis_a}}
  \end{subfigure}
  \begin{subfigure}[text]{0.24\columnwidth}
    {\includegraphics[width=0.99\columnwidth]{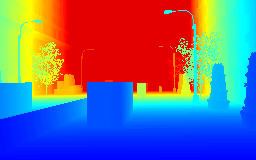}}
     {\caption{\footnotesize Depth GT}\label{fig:depth_analysis_b}}
  \end{subfigure}
  \begin{subfigure}[text]{0.24\columnwidth}
    {\includegraphics[width=0.99\columnwidth]{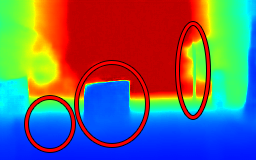}}
     {\caption{\footnotesize Depth-only}\label{fig:depth_analysis_c}}
  \end{subfigure}
  \begin{subfigure}[text]{0.24\columnwidth}
    {\includegraphics[width=0.99\columnwidth]{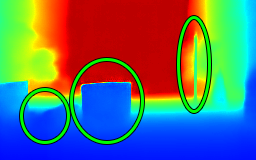}}
     {\caption{\footnotesize J-MOD$^{2}$}\label{fig:depth_analysis_d}}
  \end{subfigure}
  
    \begin{subfigure}[text]{0.24\columnwidth}
    {\includegraphics[width=0.99\columnwidth]{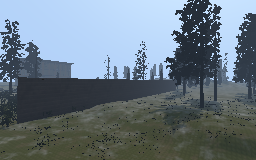}}
    {\caption{\footnotesize RGB image}\label{fig:depth_analysis_e}}
  \end{subfigure}
  \begin{subfigure}[text]{0.24\columnwidth}
    {\includegraphics[width=0.99\columnwidth]{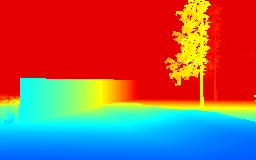}}
    {\caption{\footnotesize Depth GT}\label{fig:depth_analysis_f}}
  \end{subfigure}
  \begin{subfigure}[text]{0.24\columnwidth}
    {\includegraphics[width=0.99\columnwidth]{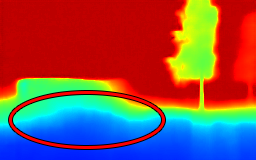}}
    {\caption{\footnotesize Depth-only}\label{fig:depth_analysis_g}}
  \end{subfigure}
  \begin{subfigure}[text]{0.24\columnwidth}
    {\includegraphics[width=0.99\columnwidth]{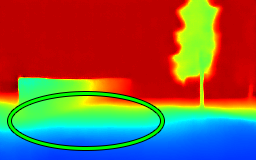}}
    {\caption{\footnotesize J-MOD$^{2}$}\label{fig:depth_analysis_h}}
  \end{subfigure}

 \caption{\small For each row, we compare J-MOD$^{2}$ depth maps with the ones predicted by the depth-only architecture. J-MOD$^{2}$ estimations are sharper and more defined. Consider, for example, the bollard and the lamppost in Figures \ref{fig:depth_analysis_a}-\ref{fig:depth_analysis_d} or the ground surface in Figures  \ref{fig:depth_analysis_e}-\ref{fig:depth_analysis_h}, whose depth is wrongly estimate by the depth-only estimator.\vspace{-2.0em}}  
        \label{fig:depth_analysis}

\end{figure}

Besides the advantages given by J-MOD$^{2}$ in terms of numerical performance, in the following, we qualitatively discuss the benefits of our joint architecture compared to its single task counterparts.

Figure \ref{fig:det_analysis} shows a comparison between the estimated obstacle bounding boxes of the detector-only architecture and the J-MOD$^{2}$ ones. It can be observed that, by exploiting the auxiliary depth estimation task, J-MOD$^{2}$ learns a detector that is aware of scene geometry. This results in an architecture that models a better concept of obstacle and, thus, is more precise in detecting what really determines a threat for the robot. Hence, it avoids wrong detections, such as ground surfaces (see Figures \ref{fig:det_analysis_a} and \ref{fig:det_analysis_b}), or full buildings of which only the closest part would constitutes an immediate danger for navigation (see Figures \ref{fig:det_analysis_c} and \ref{fig:det_analysis_d}).

Similarly, depth estimation branch of the proposed J-MOD$^{2}$ approach takes advantage from the obstacle detector task to refine the estimation of the scene geometry. The representation learned by the J-MOD$^{2}$ depth estimation stream contains also visual clues about object shapes and proportions, which gives it the capability to integrate object semantics when estimating the scene depths.
Compared to the depth-only architecture \cite{mancini2017towards}, our approach predicts sharper and more precise depth maps. This is more evident if we consider very thin elements and objects that could be mistaken for ground surfaces (\eg consider the lamppost and the bollard in Figures \ref{fig:depth_analysis_a}\ref{fig:depth_analysis_d} or the ground estimates in Figures \ref{fig:depth_analysis_e}\ref{fig:depth_analysis_h}).

\subsection{Navigation experiments}
%

\begin{figure*}[ht!]
\centering
  \ffigbox{}
  {
    \CommonHeightRow
    {
      \begin{subfloatrow}[2]
      \hspace{-2.8em}
    \ffigbox[\FBwidth]
    {\includegraphics[height=\CommonHeight]{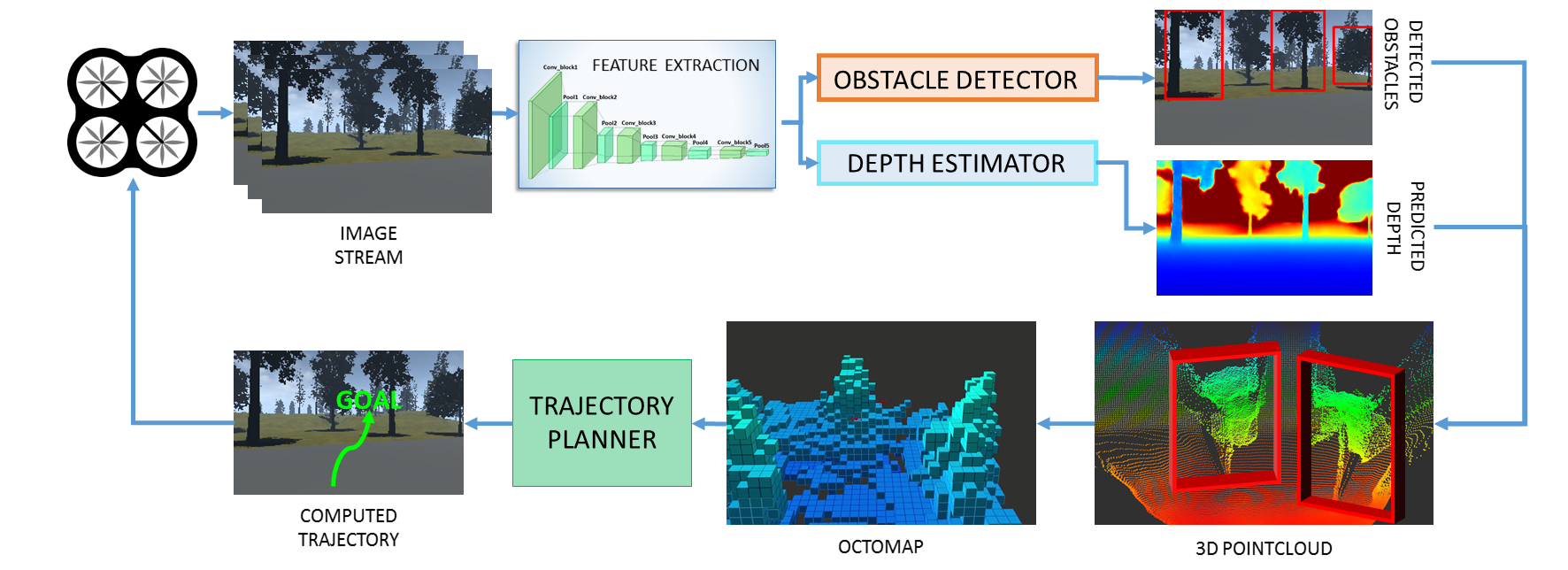}}
    {\vspace{0.5 em}}
    \hspace{0.5em}
    \ffigbox[\FBwidth]
    {\includegraphics[height=\CommonHeight]{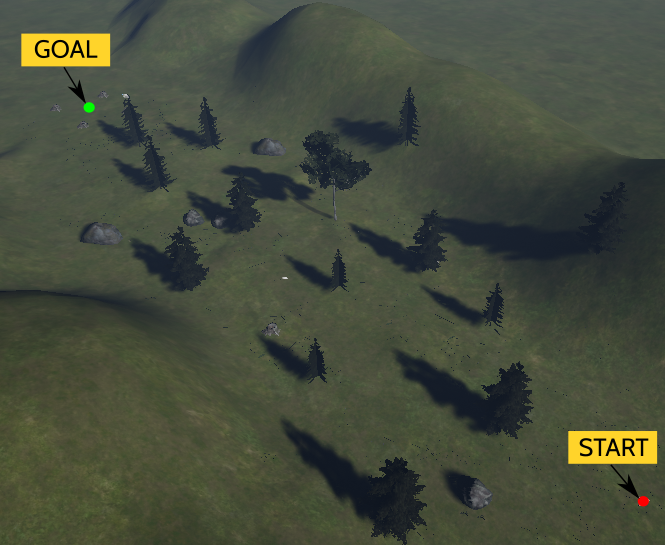}}
    {\vspace{0.5em}}
      \end{subfloatrow}
    } 
    \vspace{-1.2em}
	\caption{\small Architecture of the full navigation pipeline (on the left) and a aerial picture of the test scenario (on the right). For each RGB image captured by the MAV frontal camera, a depth map is computed and converted into a point cloud used to update the 3D map and compute an obstacle-free trajectory. The MAV then flies along the computed trajectory until a new obstacle is detected. \label{fig:navigation}}

    }
    \vspace{-1.2em}
\end{figure*}

We further validate J-MOD$^{2}$ effectivness for obstacle detection applications by setting up a simulated full MAV navigation system. We depict the system architecture in Figure \ref{fig:navigation}. We create a virtual forest scenario on Unreal Engine, slightly different from the one used for dataset collection. The line-of-sight distance between the takeoff point and the designed landing goal is about $61$ meters. Trees are about $6$ meters tall and spaced $7$ meters from each other, on average. An aerial picture of the test scenario is reported in Figure \ref{fig:navigation}. 

A simulated MAV is able to navigate into the scenario and collect RGB images from its frontal camera. We estimate depth from the captured input and we employ it to dynamically build and update an Octomap \cite{hornung2013octomap}. We plan obstacle-free trajectories exploiting an off-the shelf implementation of the RRT-Connect planner \cite{kuffner2000rrt} from the \textit{MoveIt!} ROS library, which we use to pilot the simulated MAV at a cruise speed of $1 m/s$. Trajectories are bounded to a maximum altitude of $5$ meters. As a new obstacle is detected along the planned trajectory, the MAV stops and a new trajectory is computed. The goal point is set $4$ meters above the ground. For each flight, we verify its success and measure the flight distance and duration. A flight fails if the MAV crashes or gets stuck, namely not completing its mission in a $5$ minute interval. We compare J-MOD$^{2}$ with the Eigen's baseline, both trained on the UnrealDataset. 

While planning, we add a safety padding on each Octomap obstacles. This enforces the planner to compute trajectories not too close to the detected obstacles. For each estimator, we set this value equal the average RMSE obstacle depth error on the UnrealDataset test set, as reported in Table \ref{tab:unreal_results}: $1.034$ meters for J-MOD$^{2}$, $1.854$ meters for Eigen. We refer to this value as a reliability measure of each estimator; the less accurate an estimator is, the more padding we need to guarantee safe operation.
We perform 15 flights for each depth estimator and report their results on Table \ref{tab:navigation}.

\begin{table}[ht!]
\vspace{0.8em}
  \centering
  \caption{\small Results of the navigation experiment. We compare the navigation success rate when using J-MOD$^{2}$ and Eigen's approach as obstacle detection systems. \vspace{-0.5em}}
  
  \label{tab:navigation}
  \begin{scriptsize}

  \begin{tabular}{|c||c|c|}
    \hline
     &  EIGEN \cite{eigen2015predicting} &  J-MOD$^{2}$ \\
	\hline
	 Success rate 								&  26,6\%&  \textbf{73,3\%}\\
	 Failure cases  								&  8 stuck / 3 crash &  2 stuck / 2 crash\\
	 Avg. flight time 					&  147s	&  \textbf{131s}\\
	 Std. Dev. Flight Time 		&  18.51s &  \textbf{12.88s}\\
	 Avg. flight distance				&  78m 	&  \textbf{77m} \\
	 Std. Dev. Flight Distance 	&  \textbf{4.47m} &  9.95m\\	
	\hline
  \end{tabular}
    \end{scriptsize}
\end{table}

J-MOD$^{2}$ clearly performs better in all metrics, proving that how our method is effective for monocular obstacle detection. By analyzing failure cases, for 6 times the MAV using Eigen as obstacle detector got stuck in the proximity of goal point because ground was estimated closer than its real distance, causing planner failure in finding an obstacle-free trajectory to the goal. J-MOD$^{2}$ failures are mostly related on erratic trajectory computation which caused the MAV to fly too close to obstacles, causing lateral collisions or getting stuck in proximity of tree's leaves.

\section{Conclusion and Future Work} \label{sec:conclusions}

In this work, we proposed J-MOD$^{2}$, a novel end-to-end deep architecture for joint obstacle detection and depth estimation. We demonstrated its effectiveness in detecting obstacles on synthetic and real-world datasets. We tested its robustness to appearance and camera focal length changes. Furthermore, we deployed J-MOD$^{2}$ as an obstacle detector and 3D mapping module in a full MAV navigation system and we tested it on a highly photo-realistic simulated forest scenario. We showed how J-MOD$^{2}$ dramatically improves mapping quality in a previously unknown scenario, leading to a substantial lower navigation failure rate than other SotA depth estimators. In future works, we plan to further improve robustness over appearance changes, as this is the major challenge for the effective deployment of these algorithms in practical real-world scenarios.   

\bibliographystyle{IEEEtran}
\bibliography{bibliography}

\end{document}